%% file: ms.tex
\documentclass[10pt,twocolumn,letterpaper]{article}

\usepackage{cvpr}
\usepackage{times}
\usepackage{epsfig}
\usepackage{graphicx}
\usepackage{amsmath}
\usepackage{amssymb}

\usepackage{graphicx,pstricks}
\usepackage{graphics}
\usepackage{amsmath}
\usepackage{multirow}
\usepackage{booktabs}       % professional-quality tables
\usepackage{amsfonts}       % blackboard math symbols
\usepackage{nicefrac}       % compact symbols for 1/2, etc.

\newcommand{\TT}[1]{\texttt{#1}}
\newcommand{\BF}[1]{\textbf{#1}}
\newcommand{\IT}[1]{\textit{#1}}

\newcommand{\M}[1]{\mathbf{#1}}

\usepackage[breaklinks=true,bookmarks=false]{hyperref}

\cvprfinalcopy % *** Uncomment this line for the final submission

\def\cvprPaperID{6173} % *** Enter the CVPR Paper ID here

\ifcvprfinal\fi
\begin{document}

%%%%%%%%% TITLE
\title{Building Efficient Deep Neural Networks with Unitary Group Convolutions}

\author{Ritchie Zhao
\and
Yuwei Hu
\and
Jordan Dotzel
\and
Christopher De Sa
\and
Zhiru Zhang
\and
Cornell University\\
Ithaca, NY 14853, USA\\
{\tt\small \{rz252, yh457, jad443\}@cornell.edu, cdesa@cs.cornell.edu, zhiruz@cornell.edu}\\
% For a paper whose authors are all at the same institution,
% omit the following lines up until the closing ``}''.
% Additional authors and addresses can be added with ``\and'',
% just like the second author.
% To save space, use either the email address or home page, not both
}

\maketitle
\thispagestyle{empty}

%%%%%%%%% ABSTRACT
\begin{abstract}
   We propose unitary group convolutions (UGConvs), a building block for CNNs which compose a group convolution with unitary transforms in feature space to learn a richer set of representations than group convolution alone.
   UGConvs generalize two disparate ideas in CNN architecture, channel shuffling (i.e. ShuffleNet~\cite{zhang2017shufflenet}) and block-circulant networks (i.e. CirCNN~\cite{ding2017circnn}), and provide unifying insights that lead to a deeper understanding of each technique.
   We experimentally demonstrate that dense unitary transforms can outperform channel shuffling in DNN accuracy.
   On the other hand, different dense transforms exhibit comparable accuracy performance.
   Based on these observations we propose HadaNet, a UGConv network using Hadamard transforms.
   HadaNets achieve similar accuracy to circulant networks with lower computation complexity, and better accuracy than ShuffleNets with the same number of parameters and floating-point multiplies.
\end{abstract}

%%%%%%%%% BODY TEXT
\input{sec-intro.tex}
\input{sec-related.tex}

\input{sec-ugconv.tex}
\input{sec-results.tex}
\input{sec-imagenet.tex}
\input{sec-conclusions.tex}
\subsubsection*{Acknowledgments}
This work was supported in part by the Semiconductor Research Corporation (SRC) and DARPA.
We would like to thank Prof. Yanzhi Wang (Northeastern University), Prof. Bo Yuan (Rutgers University), and their students for providing technical details and code regarding CirCNN~\cite{ding2017circnn}.

%medskip
{\small
\bibliographystyle{ieee_fullname}
\bibliography{ms}
}

\end{document}

%% file: sec-intro.tex
\section{Introduction}

Deep convolutional neural networks (CNNs) have proven extremely successful at large-scale computer vision problems. Research over the past few years has made steady progress on improving CNN accuracy~\cite{xu2018scaling}.
Concurrently, efforts have been made to reduce the number of parameters and floating-point multiplies (fpmuls) in CNNs.
One major trend in this research space is the increasing sparsity of layer connections.
Early networks such as AlexNet~\cite{krizhevsky2012imagenet} and VGG~\cite{simonyan2015vgg} exclusively utilize dense mappings, i.e. convolutional (conv) or fully-connected (FC) layers that form a weight connection between every input and every output feature.
More advanced architectures such as Xception~\cite{chollet2016xception} and MobileNets~\cite{howard2017mobilenets} make use of \IT{depthwise separable} convolutions, which consist of a sparse spatial mapping (depthwise convolution) and a dense cross-channel mapping (pointwise convolution).
Even more recently, ShuffleNet~\cite{zhang2017shufflenet} replaces the pointwise convolutions with sparse \IT{group convolutions}, and additionally proposes a channel shuffle to allow information to flow between groups.
These changes to layer structure look to remove weight connections while retaining accuracy performance.

%  Group convolutions were popularized earlier in ResNeXt~\cite{xie2017resnext}, 
A different line of efficient CNNs research looks to train networks with circulant or block-circulant~\footnote{In this paper, block-circulant, block-diagonal, etc. refers to matrices consisting of square sub-matrices which are circulant, diagonal, etc. This is different from the canonical definition of a block-diagonal matrix.} weights~\cite{cheng2015circulant, sindhwani2015structured, ding2017circnn, wang2018clstm}.
An $n \times n$ circulant matrix contains only $n$ unique elements. Moreover, every circulant matrix $\M{C}$ can be diagonalized by the normalized discrete Fourier matrix $\M{F}$ as follows:
\begin{equation}
    \label{eqn:circulant}
    \M{C} = \M{F}^* \M{D F}
\end{equation}
giving rise to an asymptotically faster algorithm for matrix multiplication via the fast Fourier transform (FFT). By exploiting these properties of circulant weights, these works can also reduce CNN complexity and model size.

In this paper, we propose the concept of \IT{unitary group convolution} (UGConv), defined as a building block for neural networks that combines a weight layer (most commonly a group convolution) with unitary transforms in feature space.
We show that group convs with channel shuffle (ShuffleNet) and block-circulant networks (CirCNN) are specific instances of UGConvs.
By unifying two different lines of work in CNN literature, we gain a deeper understanding into the basic underlying idea --- that group convolutions exhibit improved learning ability when performed in a transformed feature basis.
Through a series of experiments, we then investigate how different transforms and UGConv structures affect the learning performance.
%The addition of unitary transforms facilitates better learning of cross-channel correlations, similar to the channel shuffles in ShuffleNet.
%Indeed, ShuffleNet is a specific instance of UGConv networks (UGConvNets) as the shuffle is a unitary permutation transform.
%Moreover, circulant and block-circulant CNNs are also examples of UGConvNets: by applying Equation~\eqref{eqn:circulant} to 2D slices of the weight tensors, we show that convolutional layers with circulant weights are equivalent to group convolutions composed with discrete Fourier transforms (DFTs) in channel space.
%Similarly, block-circulant conv layers are identical to group convolutions in a transformed Fourier-like basis.
%These examples from literature lend evidence to the idea that UGConvs can outperform ordinary group convolutions. Furthermore, the use of two different transforms (DFT and permutation) motivates us to study a variety of unitary transforms.
Specifically, our contributions are as follows:
\begin{enumerate}
    \item We propose the concept of unitary group convolutions. We show that ShuffleNets and circulant networks, techniques from two disparate lines of research, are in fact both instances of UGConv networks. This lets us unify the conceptual insights of both works.
    
    \item We evaluate how different unitary transforms affect learning performance. Our experiments show that when the weight layer is highly sparse (i.e. the number of groups is large), dense transforms outperform simple permutations.
    
    \item We propose HadaNets, UGConv networks using the easy-to-compute Hadamard transform. HadaNets obtain similar accuracy as circulant networks at a lower computation complexity, and outperform ShuffleNets with identical parameter and fpmul counts.
\end{enumerate}

%% file: sec-related.tex
\section{Related Work}

\subsection{Depthwise Separable and Group Convolutions}
\label{subsec:related-group}
In a traditional convolutional layer, each 3D filter must learn both spatial and cross-channel correlations. A depthwise separable convolution decouples this into two steps: a depthwise convolution which only performs spatial filtering, and a pointwise convolution which only learns cross-channel mappings.
The idea originated in Sifre 2014~\cite{sifre2014thesis} and was subsequently popularized by networks like Xception~\cite{chollet2016xception} and MobileNets~\cite{howard2017mobilenets}.
These works showed that depthwise separable convolutions can outperform traditional convolutions using fewer parameters and fpmuls.

A group convolution divides the input and output features into mutually independent groups and performs a convolution in each one. Depthwise convs are specific cases of group convs with group size $1$.
%; group convs are also equivalent to depthwise convs with a \IT{channel multiplier} and sum reduction.
Group convolutions were part of the original AlexNet, but only to facilitate training on multiple GPUs~\cite{krizhevsky2012imagenet};
they gained popularity as a building block of efficient CNNs as part of ResNeXt~\cite{xie2017resnext} and ShuffleNet~\cite{zhang2017shufflenet}. The latter proposed channel shuffling to promote cross-channel information flow, surpassing MobileNets in accuracy and parameter efficiency.

Interleaved group convolutions~\cite{zhang2017igc,xie2018igcv2,sun2018igcv3} examines interleaving group convs and channels shuffles, and showed how a specific combination of width and sparsity (i.e. number of groups) can maximize accuracy.
Deep Roots~\cite{ioannou2017deeproots} uses group convolutions with increasing group size deeper into the network to improve numerous existing models.
Distinct from these works, we study the composition of group convs with dense unitary transforms.

%group convs were 3x3 in resnext, shufflenet is first to use 1x1 group convs
\subsection{Circulant and Block-Circulant Networks}
\label{subsec:related-circulant}
An $n$-by-$n$ circulant matrix requires only $O(n)$ storage space and $O(n \log n)$ operations for the matrix-vector product (see Equation~\eqref{eqn:circulant}). Circulant weights can reduce the model size and computational complexity of CNNs in a deterministic manner.
Cheng et al. in 2015 applies this to achieve 18x parameter reduction on AlexNet with only 0.7\% Top-1 accuracy loss~\cite{cheng2015circulant}.
Other authors proposed variations of circulant structure. Moczulski et al.'s ACDC used cosine transforms to avoid complex values that arise with DFTs and added a second channel-wise filter~\cite{moczulski2016acdc}. Sindhwani et al. studied the superset of generalized Toeplitz-like matrices~\cite{sindhwani2015structured}. These works exclusively worked on structured FC layers.

More recently, Wang et al.~\cite{ding2017circnn, wang2018circulant} proposed to use block-circulant matrices and applied them to both FC and convolutional layers. Block-circulant structure elegantly addresses the long-standing issue of non-square weight matrices.
%The same authors also proved the universal approximation property for circulant nets.
The same authors also leveraged the butterfly structure of the DFT to construct efficient accelerators for circulant nets in dedicated hardware~\cite{ding2017circnn, wang2018clstm}.
A more recent follow-up in this line of work proposed to use permuted block-diagonal matrices in specialized hardware~\cite{deng2019permdnn}.

\subsection{Random Projections and Hadamard Networks}
\label{subsec:related-random}
Our study on random orthogonal and Hadamard transforms is partly inspired by the Fastfood transform~\cite{le2013fastfood} and its application to CNNs~\cite{yang2015deepfried}. This work is a well-known example of using random embeddings and Hadamard transforms in machine learning.
%We note that the adaptive Fastfood transform falls under the umbrella of UGConvs.
%Discuss Johnson–Lindenstrauss Lemma?

A recent work from Devici et al.~\cite{deveci2018hadamard} used Hadamard-transformed images as CNN inputs. Their work differs significantly from ours; they applied a single 2D Hadamard on the input image to extract frequency features while we use Hadamard throughout the network for channel mixing.

%% file: sec-ugconv.tex
\input{figs/fig-ugconv.tex}

\section{Unitary Group Convolutions}
The basic idea of a UGConv is a group convolution sandwiched between two unitary transforms in feature space.
%Below we give a formal definition of UGConvs and show how it generalizes ShuffleNet blocks and circulant nets.
Let $\M{X}$ be an $M$-channel input tensor to a conv/FC layer. Each channel is a 2D feature map (for a dense layer the dimensions are $1 \times 1$).
Let $\M{x}^{(i)}$ denote the $i$'th channel in $\M{X}$.
Similarly, let $\M{Y}$ be the $N$-channel output tensor, and let $\M{W}$ be the weight tensor consisting of $M \times N$ filters.
We can now define an ordinary conv layer below:
\[
  %\label{eqn:conv}
  \M{y}^{(j)} = \sum_{i=1}^M \M{x}^{(i)} * \M{W}^{(ij)}, \qquad 1 \le j \le N
\]
Figure~\ref{fig:ugconv}(a) illustrates such a conv or dense layer.
Note that although the figure looks like matrix multiplication, each square represents a 2D weight filter or feature map.
%A group convolution is simply a collection of $G$ convolutions:
%\[
%  %\label{eqn:group-conv}
%  \M{y}^{(g,j)} = \sum_{i=1}^{M/G} \M{x}^{(g,i)} * \M{W}^{(g,ij)}
%\]

A group convolution is simply a collection of $G$ disjoint convolutions ($G$ is the number of groups). Each conv takes $M/G$ input channels and produces $N/G$ output channels.
\begin{equation}
    \label{eqn:gconv}
    \M{\tilde{y}}^{(g,j)} = \sum_{i=1}^{M/G} \M{\tilde{x}}^{(g,i)} * \M{\tilde{W}}^{(g,ij)}, \quad 1 \le j \le N/G\\
\end{equation}
Here $g$ denotes the group ($1 \le g \le G$), and we re-index $\M{x}$ and $\M{y}$ with two indices (\IT{group}, \IT{channel in group}). Figure~\ref{fig:ugconv}(b) illustrates how non-zero weights in a group conv form $\frac{M}{G} \times \frac{N}{G}$ blocks along the main diagonal. A group conv reduces parameter size and fpmuls by a factor of $G$ relative to an ordinary conv. However, this is achieved by removing all weight connections between groups and negatively impacts learning behavior.

A UGConv recovers this lost learning ability by sandwiching the group conv between two cross-channel unitary transforms $\M{P}$ and $\M{Q}$ (Figure~\ref{fig:arch}(a)). More formally, we can define a UGConv is:
\begin{equation}
    \begin{aligned}
      \label{eqn:ugconv}
      \M{\tilde{X}}_k &= \M{PX}_k \qquad \forall k\\
      \M{\tilde{y}}^{(g,j)} = \sum_{i=1}^{M/G} \M{\tilde{x}}^{(g,i)}& * \M{\tilde{W}}^{(g,ij)}, \quad 1 \le j \le N/G\\
      \M{Y}_l &= \M{Q\tilde{Y}}_l \qquad \forall l
    \end{aligned}
\end{equation}
For a tensor $\M{X}$ containing $M$ channels, $\M{X}_k$ is defined as the $M$-length vector formed by taking the $k$'th element/pixel from each channel. $\M{P} \in \mathbb{C}^{M \times M}$ and $\M{Q} \in \mathbb{C}^{N \times N}$ are unitary matrix transforms applied element-wise over the input and output channels.
We use tilde ($\M{\tilde{x}, \tilde{y}, \tilde{W}}$) to indicate tensors in the transformed feature space.
Note that: (1) $\M{P}$ and $\M{Q}$ can be identity transforms, and thus UGConv includes group convolutions; (2) unitary transforms preserve inner products, thus they should not diminish gradient magnitudes in the network; (3) UGConv can also be applied to FC layers (using $1 \times 1$ feature maps and a $1 \times 1$ group conv).

\input{figs/fig-arch.tex}

One key point to make is the equivalence between a \IT{group conv} and a convolution with \IT{block-diagonal weights} (i.e. weights which consist of sub-blocks of square diagonal matrices).
Figure~\ref{fig:ugconv}(c) shows a block-diagonal conv, which visually already looks identical to the group conv in Figure~\ref{fig:ugconv}(b).
More formally, divide $\M{X}$ and $\M{Y}$ into size $D \times 1$ sub-blocks, and $W$ into $D \times D$ sub-blocks which are diagonal. Let $i$ index the input sub-blocks ($0 \le i \le M/D-1$), $j$ index the output sub-blocks ($0 \le j \le N/D-1$), and $d$ index the channels within each sub-block ($1 \le d \le D$). We can express the block-diagonal conv as follows:
\begin{align*}
    \M{y}^{(j*D+d)} &= \sum_{i=0}^{M/D-1} \M{x}^{(i*D+d)} * \M{W}^{(i*D+d\,\,\, j*D+d)}
\end{align*}
Only $D$ convs need to be performed for each $D \times D$ sub-block because they are diagonal. Similar to Equation~\ref{eqn:gconv}, we can simplify notation by re-labeling using a tuple (\IT{sub-block}, \IT{channel in sub-block}). This removes the multiplies by $D$ and allows $i$ and $j$ to start from 1. Then:
\begin{equation}
    \label{eqn:diag-conv}
    \M{y}^{(j,d)} = \sum_{i=1}^{M/D} \M{x}^{(i,d)} * \M{W}^{(ij,d)},\quad 1 \le j \le N/D\\
\end{equation}
It is easy to see that Equation~\ref{eqn:diag-conv} matches Equation~\ref{eqn:gconv}.

\subsection{UGConv and ShuffleNet}
\label{subsec:ugconv-shufflenet}
ShuffleNet is a variant of the MobileNets architecture in which the pointwise convolutions (which take up $93.4\%$ of the multiply-accumulate operations~\cite{zhang2017shufflenet}) are converted into group convolutions. However, when multiple group convs are stacked together, the lack of connections between groups over many layers prevents the learning of cross-group correlations.
To address this, ShuffleNet shuffles the output channels groups in a fixed, round-robin manner --- for each group, the first channel is shuffled into group 1, the second channel into group 2, etc.
This shuffle can be expressed as a permutation in feature space, and ShuffleNets are thus an example of of UGConvNets where $\M{P}$ is identity and $\M{Q}$ is a fixed permutation matrix.

ShuffleNet shows experimentally that it is beneficial to shuffle information across groups when stacking group convs. However, shuffling channels is not the only way to accomplish such information mixing.
%However, the shuffling becomes less effective when the number of groups is large --- e.g. when each channel is in its own group, shuffling the channels around does nothing. There may be other unitary transforms that can better accomplish cross-channel mixing.

\subsection{UGConv and Circulant Networks}
\label{subsec:ugconv-circnn}
Circulant and block-circulant neural networks~\cite{ding2017circnn, wang2018circulant} utilize layers that impose a block-circulant structure on their weight tensors.
For an FC layer, the 2D weight matrix is made to be circulant.
For a conv layer, the circulant structure is applied over the input and output channels axes.
That is to say, given a 4D convolutional weight tensor with shape \TT{(height, width, in\_channels, out\_channels)}, each 2D slice of this tensor \TT{[i,j,:,:]} becomes circulant.

Figure~\ref{fig:ugconv}(d) shows a block-circulant layer where each $2 \times 2$ sub-block of the weight tensor is circulant.
By Equation~\eqref{eqn:circulant}, each $D \times D$ circulant matrix can be decomposed into a  $D$-length DFT, a diagonal matrix, and a corresponding IDFT.
In Figure~\ref{fig:ugconv}(e), each $D \times D$ sub-block is diagonalized in this fashion. We use tilde to indicate weight values in the DFT-transformed space. The resulting weight structure is block-diagonal, and the weight layer sits between two block-DFT transforms.
We know from the previous section that block-diagonal weights correspond to group convolutions.
Therefore, \IT{a block-circulant layer is just a group convolution in a transformed feature space}.
This of course falls within the definition of a UGConv, with $\M{P}$ and $\M{Q}$ being block-DFT/IDFT transforms.
Note that these DFTs are applied along the channels, and so circulant networks are \IT{not} examining the spatial frequency components of the image.

We make a few additional notes about block-circulant layers. First, the size of the circulant blocks $D$ is equal to the \IT{number of groups} in the equivalent group conv (not the group size).
Thus each $D$-length DFT touches a single channel in every group, fully mixing information between groups.
Second, though our example uses a "square" weight tensor (i.e. $M = N$), non-square block-circulant tensors can be diagonalized as well. As long as both $M$ and $N$ are divisible by $D$, the 'rectangular' weight tensor can be divided into $D \times D$ blocks.
In this case, $\M{P \in \mathbb{C}^{M \times M}}$ is not the inverse of $\M{Q \in \mathbb{C}^{N \times N}}$, but each sub-block along the diagonal of $\M{P}$ is the inverse of the corresponding sub-block in $\M{Q}$. We say that $\M{P}$ is the \IT{block-inverse} of $\M{Q}$.

Because $\M{P}$ and $\M{Q}$ are block-inverses, if we directly stack multiple such blocks many of the transforms will cancel out. However, practical DNNs include batch norm and/or nonlinearities between linear layers. The block-DFTs (and orthogonal transforms in general) do not commute with channel-wise or pointwise operations, which prevents trivial cancellation. However, note that channel shuffles \IT{do} commute and cancel out in this manner.

\subsection{Discussion of UGConvs}
\label{subsec:ugconv-discussion}
We have provided two specific examples from literature (ShuffleNet~\cite{zhang2017shufflenet} and CirCNN~\cite{ding2017circnn}) which combine a structured sparse weight layer (group convolution) with unitary transforms. The transforms help to improve cross-channel representation learning without adding additional parameters. However, the two techniques have important differences.
ShuffleNet's permutations are very lightweight as they require no arithmetic operations. However, permutations do not affect the sparsity of weight layer.
On the other hand, CirCNN composes block-DFTs with a group conv to create an effective weight structure (i.e. circulant weights) which is dense. Moreover, it does so while still having less asymptotic computational complexity than unstructured dense weights.

We hypothesize that the representation learning capability of a UGConv layer is a function of both the sparsity of the weights as well as that of the transform. An unstructured dense weight layer offers the best learning capability; grouping introduces sparsity and degrades cross-channel learning performance, some of which can be recovered via transforms.
Because dense transforms create dense weight structures (i.e. circulant weights), we believe they enable learning a richer set of representations compared to sparse transforms (i.e. channel shuffling).
When the weight sparsity is low (i.e. number of groups is small), the difference between the two may be negligible in terms of network accuracy. However, we expect dense transforms to outperform shuffling when using many groups.

Another difference is that ShuffleNet applies channel shuffle on only one side of the weight layer, while CirCNN effectively applies transformations on both sides. We use the terms 1-sided and 2-sided UGConvs to refer to these two cases, and test both in our experiments.

\input{tabs/tab-hada.tex}

\input{tabs/tab-mnist2.tex}

\subsection{The Hadamard Transform}
\label{subsec:ugconv-hadamard}
One drawback of dense transforms such as DFT is that they require more computational overhead as compared to shuffling. Even using the 'fast' algorithm, each $n \times n$ DFT requires $O(n \log n)$ floating-point multiplies and adds. Furthermore, the fact that the DFT uses complex numbers may further complicate software/hardware implementations.
Finally, the DFT is taken over the channels where there is no spatial structure --- the transform exists purely to mix information across channels and not to perform domain-specific analysis. Given this, we would like to find a more efficient alternative.

The Hadamard transform~\cite{pratt1969hadamard} is defined as a matrix containing only $+1$/$-1$ elements and whose rows and columns are mutually orthogonal.
Table~\ref{tab:hadamard} shows a $4 \times 4$ Hadamard matrix.
Because all coefficients have magnitude $1$, the transform can be computed without multiplies, i.e. using adds/subtracts only.
This is extremely important as floating-point multiplies are typically the computational bottleneck for DNN computation on both GPUs and specialized hardware.
In addition, the Hadamard transform can be generated recursively like the Fourier transform, meaning that a fast Hadamard transform (FHT) exists similar to the FFT to compute a $n$-length Hadamard transform in $O(n \log n)$ adds/subtracts~\cite{pratt1969hadamard}.
The recursive nature of FHT also enables Hadamard kernels to be implemented without explicitly storing the matrix itself; instead the matrix can be generated on the fly (similar to existing implementations of FFT kernels). This means neither FHT nor FFT requires storing additional parameters.
%Table~\ref{tab:hadamard} compares the floating-point multiplies and adds needed for the DFT and Hadamard transforms, showing how the latter requires far fewer fpmuls.
%Further discussion on using the Hadamard transform in DNNs can be found in Section~\ref{subsec:practicality}.

Hadamard is more efficient that DFT, but does it achieve the same learning performance? There is some high-level intuition that this would be the case: Table~\ref{tab:hadamard} compares the weight structure imposed by $\M{P^*DP}$ when $\M{P}$ is DFT and Hadamard. DFT results in a circulant matrix; Hadamard results in a nearly identical weight matrix with only a few different elements. We hypothesize that there will be no accuracy impact in replacing circulant weights with Hadamard-diagonalizable weights in neural nets.

We further speculate that dense unitary transforms in general, including DFT and Hadamard, achieve comparable learning performance. This is again because the ordering of channels in DNNs is essentially random (i.e. the channel order encodes no useful information), meaning there are no patterns that can be exploited by one particular cross-channel transform and not others.
The transforms in UGConv exist solely to connect different channel groups, and any dense transform will work as well as another. To test this hypothesis, we experiment with randomly generated orthogonal transforms in addition to DFT and Hadamard.

%Because these operations are adds/subs, the FHT is much faster than the FFT in specialized hardware such as ASICs and FPGAs.
%\fixme{(zz: I still think we should include the table from the FB proposal. CDS: If we have space we should, otherwise not a priority.)}

%One drawback of Hadamard transforms is that they have only been proven to exist for power-of-two (and some specific even-numbered) lengths~\cite{}, while the DFT exists for all even lengths.
%This restricts the group size in a Hadamard UGConv to powers of two. However, we believe this is not a serious issue in practice as power-of-two widths and group sizes are extremely common~\cite{he2015resnet, xie2017resnext, howard2017mobilenets, chollet2016xception, simonyan2015vgg}.

%% file: figs/fig-ugconv.tex
\begin{figure*}[t]
  \centering
  \begin{minipage}{0.42\textwidth}
    \centering
    \includegraphics[width=\textwidth]{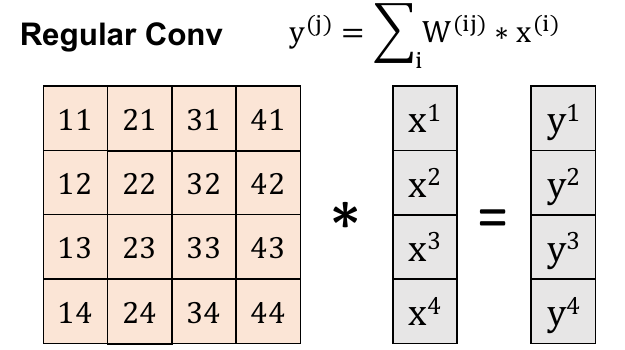}
  \end{minipage}
  \Large{(a)}
  \hspace{8px}
  %\vline width 1pt
  \hspace{4px}
  \begin{minipage}{0.42\textwidth}
    \centering
    \includegraphics[width=\textwidth]{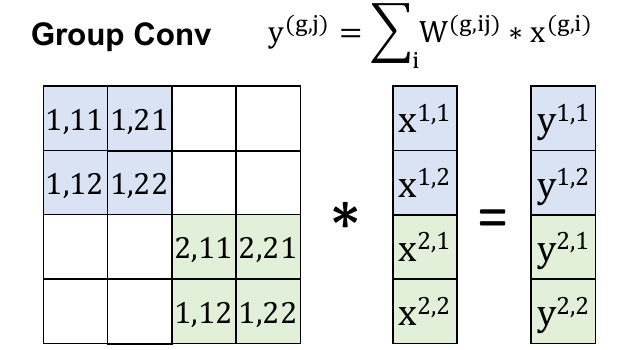}
  \end{minipage}
  \hspace{2px}
  \Large{(b)}
  \hfill

  \vspace{4px}
  
  \rule{\textwidth}{1pt}
  
  \vspace{4px}
  
  \centering
  \begin{minipage}{0.42\textwidth}
    \centering
    \includegraphics[width=\textwidth]{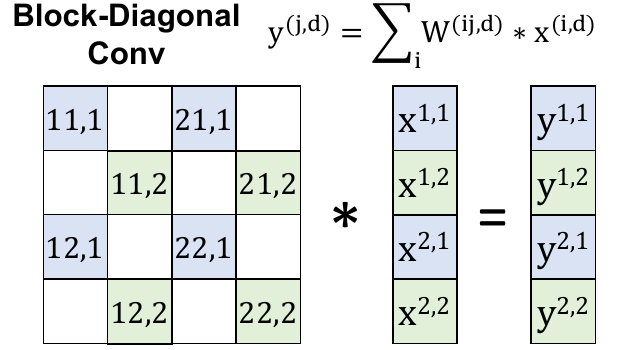}
  \end{minipage}
  \Large{(c)}
  \hspace{8px}
  %\vline width 1pt
  \hspace{4px}
  \begin{minipage}{0.42\textwidth}
    \centering
    \includegraphics[width=\textwidth]{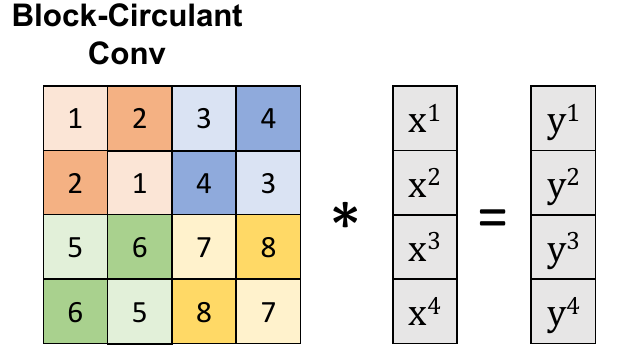}
  \end{minipage}
  \hspace{2px}
  \Large{(d)}
  \hfill

  \vspace{4px}
  
  \rule{\textwidth}{1pt}
  
  \vspace{4px}
  
  \centering
  \begin{minipage}{0.75\textwidth}
    \centering
    \includegraphics[width=\textwidth]{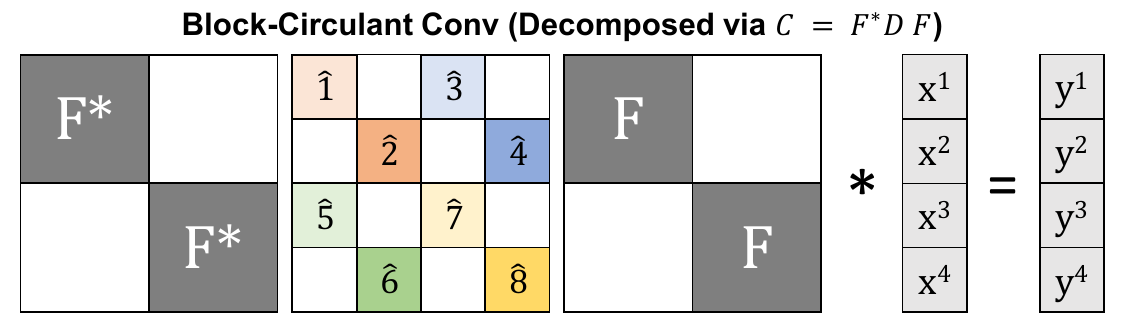}
  \end{minipage}
  \hspace{2px}
  \Large{(e)}

  \vspace{8px}

  \caption{\BF{Relation between group convs and circulant weights --} each square represents a 2D feature/filter, which can be $1 \times 1$ for an FC layer. (a) regular conv layer; (b) group conv with 2 groups; (c) same group conv reordered to show the block-diagonal weight structure; (d) block-circulant conv layer; (e) same block-circulant conv decomposed into block-DFTs and a block-diagonal conv layer.
  }
  \label{fig:ugconv}
  \vspace{-10px}
\end{figure*}

%% file: figs/fig-arch.tex
\begin{figure*}[t!]
  \hfill
  \begin{minipage}{0.25\textwidth}
    \centering
    \vspace{10pt}
    
    \includegraphics[width=\textwidth]{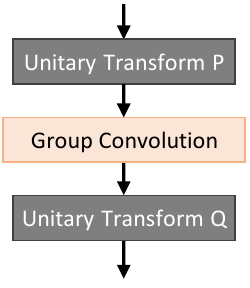}
    
    \vspace{10pt}
    (a)
  \end{minipage}
  \hfill
  \begin{minipage}{0.2\textwidth}
    \centering
    \includegraphics[width=\textwidth]{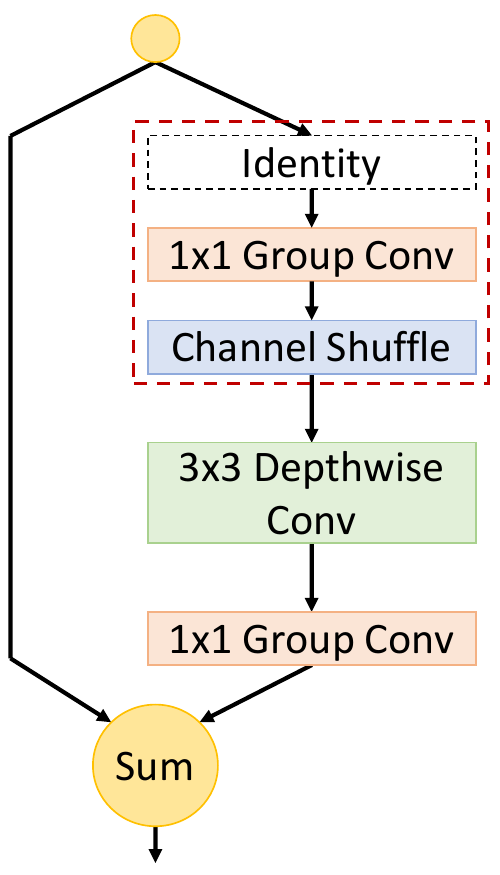}
    
    (b)\hspace{0.36\textwidth}
  \end{minipage}
  \hfill
  \begin{minipage}{0.2\textwidth}
    \centering
    \includegraphics[width=\textwidth]{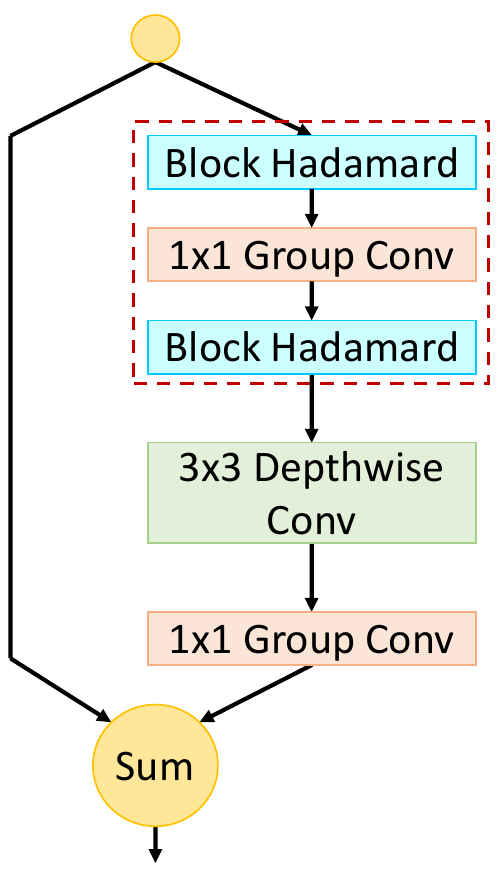}
    
    (c)\hspace{0.36\textwidth}
  \end{minipage}
  \hfill
  \vspace{8pt}
  
  \caption{\BF{CNN block architectures --} (a) a general block for unitary group convolutions; (b) a ShuffleNet block reproduced from the original paper~\cite{zhang2017shufflenet}; (c) our proposed HadaNet variation. Note that both ShuffleNet and HadaNet blocks contain the UGConv pattern.
  }
  \label{fig:arch}
\end{figure*}

%% file: tabs/tab-hada.tex
\begin{table}[tbp]
  \centering
  \caption{\BF{Hadamard vs. Discrete Fourier transforms --} The entries of the DFT matrix are the complex roots of unity. The entries of the Hadamard matrix are $+1$ or $-1$. The last column shows the structure of $\M{P}^*\M{DP}$ where $\M{D}$ is a diagonal matrix and $\M{P}$ is the transform; differences are bolded.
  }
    \begin{tabular}{c|c|c}
    \toprule
    & \BF{Fourier} & \BF{Hadamard} \\
    \toprule
    \shortstack{Transform\\$\M{P}$} & \small{$\begin{bmatrix} 1 & 1 & 1 & 1 \\ 1 & \omega & \omega^2 & \omega^3 \\ 1 & \omega^2 & \omega^4 & \omega^6 \\ 1 & \omega^3 & \omega^6 & \omega^9 \end{bmatrix}$} & \small{$\begin{bmatrix} 1 & 1 & 1 & 1 \\ 1 & -1 & 1 & -1 \\ 1 & 1 & -1 & -1 \\ 1 & -1 & -1 & 1 \end{bmatrix}$} \\
    \midrule
    \shortstack{Structure\\of $\M{P*DP}$} & $\begin{bmatrix} a & b & c & d \\ d & a & b & c \\ c & d & a & b \\ b & c & d & a \end{bmatrix}$ & $\begin{bmatrix} a & b & c & d \\ \BF{b} & a & \BF{d} & c \\ c & d & a & b \\ \BF{d} & c & \BF{b} & a \end{bmatrix}$ \\
    \midrule
    FP Muls & $n \log n$ & 0 \\
    \midrule
    FP Adds & $n \log n$ & $n \log n$\\
    %\vspace{2pt}
    \bottomrule
    \end{tabular}%
    \label{tab:hadamard}%
    \vspace{-5px}
\end{table}

%% file: tabs/tab-mnist2.tex
% Table generated by Excel2LaTeX from sheet 'mnist_arXiv'
\begin{table*}[t!]
  \centering
  \caption{\BF{Test error on a toy MNIST network --} a 'G' in the layer width columns indicates a group layer. In the transform columns, $\M{P}$ and $\M{Q}$ denote 1-sided pre-conv and post-conv transforms, respectively; $\M{PQ}$ denotes a 2-sided transform. All values are averaged over 5 runs and 90\% confidence bounds for each value are at most $\pm5\%$.}
    \begin{tabular}{lll|c|ccc|ccc}
    \toprule
    \multicolumn{3}{c|}{\BF{Layer Width}} & \multicolumn{7}{c}{\BF{Transform}} \\
    \midrule
          \BF{L2} & \BF{L3} & \BF{L4} & \BF{None} & \multicolumn{3}{c}{\BF{Rand Ortho}} & \multicolumn{3}{|c}{\BF{Rand Perm}} \\
          Conv3x3 & FC & FC &       & P & Q & PQ & P & Q & PQ \\
    \midrule
    \midrule
    20 & 20,G & 10 & 6\%     & \BF{4\%}     & \BF{4\%}     & \BF{4\%}     & 5\%     & 6\%     & 5\% \\
    \midrule
    20 & 20,G & 10,G & 27\%    & 10\%    & 8\%     & \BF{4\%}     & 27\%    & 26\%    & 25\% \\
    \midrule
    20,G & 20,G & 10 & 25\%    & \BF{10\%}    & \BF{10\%}    & \BF{10\%}    & 27\%    & 20\%    & 21\% \\
    \midrule
    20,G & 20,G & 10,G & 60\%    & 23\%    & \BF{17\%}    & 20\%    & 57\%    & 55\%    & 57\% \\
    \bottomrule
    \end{tabular}%
  \label{tab:mnist}
  %\vspace{-5px}
\end{table*}%

%% file: sec-results.tex
\section{Experimental Validation}
\label{sec:experiment}

We first present ablation studies on a toy MNIST network followed by deeper CIFAR-10 models. These experiments build up insights on UGConv.
We then demonstrate the utility of Hadamard using grouped ResNets and a ShuffleNet model from literature trained on ImageNet.

\subsection{Dense Transforms vs. Shuffle}
Our first experiment uses a toy MNIST network. This allows us to isolate the UGConv block and to compare dense orthogonal transforms versus permutations in a simple setting. We stress that the goal here is \IT{not} to build a realistic classifier. 
The layer architecture is denoted below, where each layer is described as (\IT{number of channels})(\IT{layer type}):
\[
    10\text{Conv}3\text{x}3-20\text{Conv}3\text{x}3-\BF{20\text{FC}}-10\text{FC}
\]
We perform $2\times2$ max pooling before each $3\times3$ conv layer, and a global average pool before the first FC layer. Each layer is followed by batch normalization and ReLU.

We convert the first FC layer of the network (20FC1, shown in bold) into a UGConv block (i.e. it becomes a grouped FC with transforms). The group number is equal to the number of channels to maximize sparsity.
From this base architecture we derive three variations: (1) convert the preceding Conv3x3 layer into group conv; (2) convert the following FC layer into group FC; (3) convert both surrounding layers into group layers. These test the performance of transforms in the context of stacked group layers. Two types of transform are evaluated: randomly generated dense orthogonal and random permutation transforms. We test with both 1-sided (using one of $\M{P}$ or $\M{Q}$ and setting the other to identity) and 2-sided UGConvs ($\M{P} = \M{Q}^{-1}$). All results are averaged over five runs, and we regenerate the random transformation matrices between runs.

Table~\ref{tab:mnist} shows our results. Due to the small size of the network, the 90\% confidence bound for these values can be as large as $\pm5\%$. Nevertheless, differences between transforms are clearly demonstrated.
When L3 is the only grouped layer in the network (row 1), transforms have little to no effect.
However, when two or more group layers are stacked together, the dense orthogonal transforms achieve improved accuracy. Permutations did not improve accuracy in any experiment.
This is a clear (albeit artificial) demonstration that when the number of groups is very large, dense transforms outperform permutations in learning ability.

Another interesting observation is that there is little difference between 1-sided and 2-sided transforms, regardless of whether the UGConv block is stacked before or after another group layer.
For example, in Table~\ref{tab:mnist} row 3, a dense orthogonal transform improves accuracy even when it is placed \IT{after} both group layers.
It may be surprising that a transform affects layers preceding it. But keep in mind that the transform also affects gradients on the backwards pass, allowing the same weights to 'see' more downstream activations during backpropagation.
Alternatively, we can view the UGConv layer as a learnable structured weight layer (see Section~\ref{subsec:ugconv-circnn}) --- within this perspective, the weight structure is a function of transforms both before or after.

\input{tabs/tab-cifar-resnet2.tex}

\subsection{Evaluation of Different Transforms}
We have shown that dense orthogonal transforms can improve over shuffles in small DNNs with large group sizes. 
To validate our results on more realistic architectures, we perform experiments on CIFAR-10~\cite{krizhevsky2009cifar} using ResNet~\cite{he2015resnet}. We use UGConvs to replace the two $3 \times 3$ convolutions in each ResNet block, and to replace the $1 \times 1$ projection layers. ResNets are divided into three stages (S1, S2, S3), with later stages having more channels.
We use more groups in later stages, keeping the ratio of channels to groups constant. Two models are tested: \BF{ResNet-20} (3 block per stage) and \BF{ResNet-56} (9 blocks per stage).
We also experiment with the same high-level architecture but using the building block from ShuffleNet~\cite{zhang2017shufflenet}. This block which contains two $1 \times 1$ convs and a $3 \times 3$ depthwise conv (see Figure~\ref{fig:arch}(b)). Following ShuffleNet we apply transformations around the first $1 \times 1$ group conv only and make no changes to the second group conv. Again, two models are tested: \BF{ShuffleNet-29} (3 block per stage) and \BF{ShuffleNet-56} (6 blocks per stage).

We use layer widths and training hyperparameters from \cite{he2015resnet} and make use of standard data augmentations: padding 8 pixels on each side and randomly cropping back to original size, combined with a random horizontal flip~\cite{he2015resnet, huang2016stochastic, lin2013nin}. Each network is trained for 200 epochs, and we report the mean test error over the last 5 epochs.% and the baseline achieves a test error rate of $7.0\%$.

We test the following transforms: identity (None), ShuffleNet permutation (Shuffle), block-Hadamard (Hada), block-DFT (Fourier), and block-random-orthogonal (Ortho). The block transforms follow the same structure described in Section~\ref{subsec:ugconv-circnn}. For each transform, both 1-sided (letting $\M{Q}$ be the transform and $\M{P}$ identity) and 2-sided ($\M{P}$ and $\M{Q}$ are block-inverses) versions are tested where reasonable.
The 1-sided DFT is left out because it introduced complex numbers into the network.
For the 2-sided channel shuffle (Shuffle*), we set $\M{P} = \M{Q}$ to essentially perform additional shuffling; this is done since using block-inverse shuffles will lead to trivial cancellation.
All results are displayed in Table~\ref{tab:cifar-resnet} --- the error rate with no transforms is given first followed by the accuracy improvement achieved with each UGConv setup. Our base error rates are high for CIFAR-10 because group convolutions significantly compress the network

A key result here is that dense orthogonal transforms perform similarly in accuracy. Fourier, Hada, and Ortho obtain results which are within a spread of 0.4\% in both 1-sided and 2-sided settings. On the other hand, the shuffle transforms (1 and 2-sided) clearly perform worse for the larger group sizes. This confirms our hypothesis that Hadamard is comparable to DFT in learning performance while being much easier to compute. It also provides evidence that \IT{all} dense UGConvs achieve comparable learning performance.

Another observation is that 2-sided transforms significantly outperform their 1-sided variants, which is different from the MNIST data.
We currently do not have an explanation for this effect. One speculation was that 2-sided transforms perform better when the number of input and output channels did not match. However, further testing with the small MNIST network showed that this was not the case.

Finally, note that the accuracy trends remained the same whether the transforms were applied to $3 \times 3$ group convs in ResNet or $1 \times 1$ group convs in ShuffleNet. This is evidence that spatial and cross-channel dependencies are effectively decoupled in convolutional layers, and that the size of the filter does not significantly affect channel-space transforms.

\iffalse
$y = xD$\\
$\begin{bmatrix}y_1 & y_2\end{bmatrix} = \begin{bmatrix}x_1 & x_2\end{bmatrix}\begin{bmatrix}w_1 & 0 \\ 0 & w_2\end{bmatrix} = \begin{bmatrix}w_1 x_1 & w_2 x_2\end{bmatrix}$\\

$y = xHD$\\
$\begin{bmatrix}y_1 & y_2\end{bmatrix} = \begin{bmatrix}x_1 & x_2\end{bmatrix}\begin{bmatrix}w_1 & w_2 \\ w_1 & -w_2\end{bmatrix} = \begin{bmatrix}(x_1 + x_2)w_1 & (x_1 - x_2)w_2 \end{bmatrix}$\\

$y = xDH$\\
$\begin{bmatrix}y_1 & y_2\end{bmatrix} = \begin{bmatrix}x_1 & x_2\end{bmatrix}\begin{bmatrix}w_1 & w_1 \\ w_2 & -w_2\end{bmatrix} = \begin{bmatrix}x_1w_1 + x_2w_2 & x_1w_1 - x_2w_2\end{bmatrix}$\\

$y = xHDH$\\
$\begin{bmatrix}y_1 & y_2\end{bmatrix} = \begin{bmatrix}x_1 & x_2\end{bmatrix}\begin{bmatrix}w_1 & w_2 \\ w_2 & w_1\end{bmatrix} = \begin{bmatrix}x_1w_1 + x_2w_2 & x_1w_2 + x_2w_1\end{bmatrix}$\\
\fi

%% file: tabs/tab-cifar-resnet2.tex
% Table generated by Excel2LaTeX from sheet 'cifar10_final'
\begin{table*}[t!]
  \centering
  \caption{\BF{Test error for UGConvs on CIFAR-10 --}
  The first three columns show the number of groups used in the three stages (S1-S3). The \BF{Base} column shows the test error with no transforms, and the other columns show \IT{improvement} in test error over this baseline. Some entries are blank due to insufficient time to complete the experiments.
  }
    \begin{tabular}{l|ccc|c|ccc|cccc|c}
    \toprule
    & \multicolumn{3}{c|}{\textbf{\# of Groups}} & \BF{Base} & \multicolumn{3}{c|}{\textbf{1-sided Transforms}} & \multicolumn{4}{c|}{\textbf{2-sided Transforms}} & \BF{Params} \\
    \midrule
    %&      &      &      &       &         &       & Rand. &         &         &      & Rand. \\
    & S1   & S2   & S3   &   & Shuffle & Hada & Ortho   & Shuffle* & Fourier & Hada & Ortho & \\
    \midrule
    \midrule
    ResNet-20
    & 4    & 8    & 16   & 19.5\% & 3.3\%  & 4.0\%  & 4.0\%   & 3.1\%  & 4.1\%  & \BF{4.2\%}  & 3.8\%   & 25K \\
    & 8    & 16   & 32   & 23.8\% & 2.9\%  & 4.3\%  & 3.9\%   & 4.1\%  & \BF{5.4\%}  & \BF{5.4\%}  & 5.3\%  & 14K \\
    \midrule
    ResNet-56
    & 4    & 8    & 16   & 16.0\% & 4.0\% & 4.4\%   & 4.2\%   & 4.0\%  & \BF{4.7\%}  & 4.5\%  & 4.6\%  & 76K \\
    & 8    & 16   & 32   & 20.6\% & 5.4\% & 6.1\%   & 6.4\%   & 5.8\%  & 7.1\%  & \BF{7.2\%}  & 6.8\%  & 41K \\
    \midrule
    ShuffleNet-29
    & 4    & 8    & 16   & 18.3\% & 2.7\%  & 2.4\%  & 3.1\%   & 3.8\%  & \BF{4.9\%}  & 4.5\% & 4.2\% & 23K\\
    & 8    & 16   & 32   & 22.1\% & 0.6\%  & 3.4\%  & 3.6\%   & 3.8\%  & 5.1\%  & 5.0\%  & \BF{5.3\%}   & 17K\\
    \midrule
    ShuffleNet-56
    & 4    & 8    & 16   & 16.2\% & 3.6\% & 3.5\%   & 3.4\%   & 3.9\%  & 4.6\%  & 4.5\%  & \BF{4.7\%} & 41K \\
    & 8    & 16   & 32   & 19.7\% & 4.3\% & 4.4\%   & 4.9\%   & 5.2\%  & \BF{6.0\%}  & \BF{6.0\%}  & \BF{6.0\%} & 29K \\
    \midrule
    \midrule
    \BF{Mean}
    & 4    & 8    & 16   & 17.5\% & 3.4\% & 3.6\%   & 3.7\%   & 3.7\%  & \BF{4.6\%}  & 4.4\%  & 4.3\% \\
    & 8    & 16   & 32   & 21.5\% & 3.3\% & 4.6\%   & 4.7\%   & 4.7\%  & \BF{5.9\%}  & \BF{5.9\%}  & \BF{5.9}\% \\
    %\midrule
    %\multicolumn{9}{l}{Baslines} \\
    %\midrule
    %\multicolumn{3}{l|}{ResNet-50} & \multicolumn{4}{l|}{0.0\%}     & 100\% & 100\% \\
    %\multicolumn{3}{l|}{ResNet-32} & \multicolumn{4}{l|}{0.8\%}     & 30\% & 37\% \\
    %\multicolumn{3}{l|}{ResNet-20} & \multicolumn{4}{l|}{1.5\%}     & 62\% & 62\% \\
    \bottomrule
    \end{tabular}%
  \label{tab:cifar-resnet}%
  \vspace{-5pt}
\end{table*}%

%% file: sec-imagenet.tex
\subsection{Hadamard Networks on ImageNet}

\input{tabs/tab-imagenet.tex}

The data from previous sections point to two regimes: at low weight sparsity (i.e. small group numbers) a simple shuffle is sufficient to maximize accuracy. At large group numbers, however, dense transforms outperform shuffles. This section evaluates the 2-sided block-Hadamard transform against shuffle on ImageNet.
Hadamard was chosen as it is far more efficient than other dense unitary transforms (see Section~\ref{subsec:ugconv-hadamard}, and ShuffleNet was used for comparison as it is highly related work and a strong baseline.
We refer to networks using Hadamard UGConvs as HadaNets.
Figure~\ref{fig:arch} compares the residual blocks of ShuffleNet and HadaNet.

Due to hardware constraints, we chose small models with fairly large group size --- this is the setting where dense transforms should perform the best compared to shuffle.
We evaluate ResNet-18 following the ImageNet architecture from~\cite{he2015resnet} and using group sizes 8 and 16 throughout the network.
We also test with the ShuffleNet-x0.25 g8, which is the smallest ShuffleNet variant from~\cite{zhang2017shufflenet}. This network has 50 layers and also uses 8 groups.
Each network was trained with the hyperparameters and learning rate schedule described in their respective papers.
We compare 1-sided shuffle to 2-sided block-Hadamard (note that ShuffleNet from literature already contains the 1-sided shuffle).
All results are displayed in Table~\ref{tab:imagenet}.
Our reproduction of ShuffleNet-x0.25-g8 achieved a Top-1 error of $53.6\%$, which is close to the $52.7\%$ reported in Table 2 of~\cite{zhang2017shufflenet}.

The results demonstrate that the Hadamard transform can indeed outperform shuffling in terms of accuracy on large scale datasets.
ResNet-18 with group convs is a non-standard model, but it serves to show that the trends observed in CIFAR-10 ResNets carry over to ImageNet.
On the other hand, ShuffleNet is a well-optimized baseline which obtains good accuracy performance on a very tight parameter and fpmul budget. In addition, despite very little hyperparameter tuning, HadaNet was able to improve slightly over ShuffleNet.

\subsection{Practicality of HadaNet}
\label{subsec:practicality}
HadaNet slightly outperforms ShuffleNet on accuracy, but requires extra floating-point adds. An $N$-channel group conv with $B$ groups requires $N^2/B$ fpmuls for the weight layer and $2N \log B$ adds for the two block-Hadamard transforms.
Compared to multiplies, additions are already much cheaper in hardware.
The last column of Table~\ref{tab:imagenet} shows the number of additions needed for each network if the fast Hadamard transform is used. The relative overhead of HadaNet is fairly small: the extra adds amount to only $2$-$5\%$ of existing multiply-accumulates in those networks.

However, the overhead of the Hadamard transform depends on a well-optimized implementation.
The reason we did not show runtime on GPU is that an $O(n \log n)$ fast Hadamard kernel operating along the channels is not currently available --- as a result our own HadaNet implementation is fairly slow.

On the other hand, we believe the Hadamard transform might be useful for specialized DNN accelerators implemented with FPGAs~\cite{microsoft2018brainwave} or ASICs~\cite{google2017tpu}.
Top computer hardware conferences already contain works demonstrating the use of circulant matrices for DNN compression in dedicated hardware~\cite{ding2017circnn, wang2018clstm, deng2019permdnn}.
These works show that DFTs can be very efficiently implemented in a dedicated module due to its recursive nature. We choose Hadamard because it also has the same recursive properties, meaning it should be even simpler in hardware due to lower computational complexity.
All-in-all, this paper reveals that in high weight sparsity regimes, dense transforms outperform simple shuffling. HadaNet is more efficient than the existing state-of-the-art dense transform (i.e. DFT transforms) while achieving similar accuracy performance in DNNs.

%% file: tabs/tab-imagenet.tex
% Table generated by Excel2LaTeX from sheet 'arXiv'
\begin{table*}[t]
  \centering
  \caption{\BF{Top-1 classification error on ImageNet --} we include data on both the original ShuffleNet (with our own code) and our pre-activation variation. Our baseline ShuffleNet implementation is close to the literature results (52.7\%). For each model we show the number of parameters and fpmuls, as well as the overhead in additions from the Hadamard transform.}
    \begin{tabular}{l|ccc|cc|c}
    \toprule
          & \textbf{Shuffle} & \textbf{Hada} & \BF{Delta} &\BF{Params} & \BF{FPmuls} & \BF{Hada Adds} \\
   \midrule
    ResNet-18 g8                    & 46.4\% & \BF{44.6\%}   & (-1.8\%) & 1.9M & 330M & 7.8M \\
    \midrule
    ResNet-18 g16                    & 55.8\% & \BF{52.3\%}   & (-3.5\%) & 1.2M & 226M & 10.4M \\
    \midrule
    %ResNet-34 g16                    & 52.1\% & \BF{45.0\%}   & (-7.1\%) & 1.9M & 343M & 15.5M \\
    %\midrule
    %ShuffleNet-x0.25 g8 (Our impl.) & 58.1\% & \BF{57.2\%}   & (-0.9\%) & 0.46M & 17M & 0.95M \\
    %\midrule
    ShuffleNet-x0.25 g8  & 53.6\% & \BF{52.6\%}   & (-1.0\%) & 0.46M & 17M & 0.95M \\
    \bottomrule
    \end{tabular}%
  \label{tab:imagenet}%
  %\vspace{-5px}
\end{table*}%

%% file: sec-conclusions.tex
\section{Conclusions and Future Work}

We introduce the concept of unitary group convolutions, a composition of group convolutions with unitary transforms in feature space. We use the UGConv framework to unify two disparate ideas in CNN literature, ShuffleNets and block-circulant networks, and provide valuable insights into both techniques.
UGConvs with dense unitary transforms demonstrate superior ability to learn cross-channel mappings versus ordinary and shuffled group convolutions.
Based on these these observations we propose HadaNet, a variant of ShuffleNet that improves accuracy on the ImageNet dataset  without incurring additional parameters or floating-point multiplies.

One future work is to replace the Hadamard transform with a trained ${0, +1, -1}$ transform; training may allow the transform to adapt to the weights, and introducing zeros enables sparse compute reduction.